\documentclass{article}
\usepackage[T2A]{fontenc}   
\usepackage[english]{babel}
\usepackage{graphicx}
\usepackage{url}
\usepackage{hyperref}
\usepackage{float}
\usepackage{amsmath} 
\usepackage{booktabs}
\usepackage{pifont}
\usepackage{orcidlink}
\usepackage{arxiv}

\renewcommand{\shorttitle}{Application of PINNs for Solving the Inverse Advection-Diffusion Problem}

\title{Application of Physics-Informed Neural Networks for Solving the Inverse Advection-Diffusion Problem to Localize Pollution Sources}
\author{
    Ivan Chuprov\orcidlink{0000-0002-1091-4442} \and 
    Denis Derkach \orcidlink{0000-0001-5871-0628} \and 
    Dmitry Efremenko \and 
    Aleksei Kychkin
}
\date{}

\newcommand{\affiliation}{%
    \footnotesize{National Research University Higher School of Economics, Moscow, Russia}
    \vspace{1em} 
}

\begin{document}

\maketitle

\vspace{-2em} 
\begin{center}
    \affiliation
\end{center}

\begin{abstract}
This paper investigates the application of Physics-Informed Neural Networks (PINNs) for solving the inverse advection-diffusion problem to localize pollution sources. The study focuses on optimizing neural network architectures to accurately model pollutant dispersion dynamics under diverse conditions, including scenarios with weak and strong winds and multiple pollution sources. Various PINN configurations are evaluated, showing the strong dependence of solution accuracy on hyperparameter selection. Recommendations for efficient PINN configurations are provided based on these comparisons. The approach is tested across multiple scenarios and validated using real-world data that accounts for atmospheric variability. The results demonstrate that the proposed methodology achieves high accuracy in source localization, showcasing the stability and potential of PINNs for addressing environmental monitoring and pollution management challenges under complex weather conditions.
\end{abstract}


\section{Introduction}
Weather forecasting and solving related challenges represent some of the most complex areas in meteorology. Numerical models, such as global circulation models \cite{Mechoso2015}, require significant computational resources to accurately describe atmospheric processes. However, these approaches often struggle with predicting local weather parameters, particularly in regions with limited observational data or sparse meteorological station networks. This limitation is especially critical when describing the migration of pollution in the atmosphere, where accurate modeling is essential for understanding pollutant transport and devising effective air quality management strategies \cite{Cieslik1991}.

Currently, numerous technology companies are leading the development of local and regional air quality monitoring systems. In industrial emission monitoring, well-established solutions exist that enable automatic control directly at the emission source, such as installations on smokestacks. However, several challenges remain unresolved, including predictive modeling and continuous analysis of pollutant dispersion, identification of emission sources and causes of maximum allowable concentration (MAC) exceedances, as well as proactive monitoring and reduction of harmful industrial emissions into the atmosphere.

In recent years, machine learning (ML) methods have emerged as promising alternatives to numerical weather prediction \cite{Waqas2024}. Compared to conventional approaches, ML techniques capture complex nonlinear relationships in data, which can enhance forecasting accuracy. However, a major limitation of ML models is their dependency on large volumes of high-quality training data. This challenge is particularly pronounced in regions with sparse meteorological station networks or when predicting rare events.

A relatively new and innovative approach, Physics-Informed Neural Networks (PINNs), addresses this limitation by incorporating physical laws directly into the model training process \cite{Goswami2023}. Unlike traditional neural networks, PINNs embed fundamental principles, such as mass, energy, and momentum conservation, into their loss functions. This allows them to use mathematical representations of physical laws, making them more reliable and accurate. 

PINNs have already been successfully applied in fields such as optics, fluid dynamics, and material science, where they have proven effective in solving both forward and inverse problems. For example, they have been used to address the inverse Navier-Stokes problems \cite{raissi2019physics}, groundwater modeling \cite{CUOMO2023106}, and advection-diffusion equations \cite{Mamud2022}, although the latter focused on water bodies rather than atmospheric processes. The ability of PINNs to generalize across parameter ranges has been particularly demonstrated in nonlinear Schr\"{o}dinger equations \cite{Chuprov2024}, showing their potential for efficient modeling and prediction across a broad spectrum of conditions.

Despite these achievements, the application of PINNs in atmospheric physics, particularly to localize sources of pollution, remains under-explored. The localization of the pollution source and the modeling of the dispersion of pollutants are critical to the development of strategies to improve air quality and mitigate the environmental impact of anthropogenic activities. PINNs, by integrating data-driven learning with physical modeling, offer a unique opportunity to tackle this challenge more effectively, even in scenarios with limited observational data.

This study focuses on using PINNs to address challenges related to atmospheric pollution, specifically in modeling pollutant dispersion and identifying pollution sources. By exploring the capabilities of PINNs in this context, we aim to develop a computational framework that accurately retrieves the coordinates of pollution sources using ground-based measurements, providing a robust tool for air quality management and environmental monitoring.

\section{Theory}
\subsection{Problem formulation}

The modeling of atmospheric pollution involves understanding the dispersion of pollutants and identifying their sources. This process is governed by the advection-diffusion equation, which describes the transport of pollutant concentration $c$ in the atmosphere:

\begin{equation}
    \frac{\partial c}{\partial t} + \mathbf{u} \cdot \nabla c - \nabla \cdot (k \nabla c) = s, \label{eq1}
\end{equation}
where $c$ is the pollutant concentration, $\mathbf{u}$ is the wind velocity vector, $k$ is the diffusion coefficient,
while $s$ is the pollutant source term.
For computational simplicity, we assume that the wind velocity field $\mathbf{u}$ is a known input, obtained either through observations or as the solution to the Navier-Stokes equations. The solution of the Navier-Stokes equations using PINNs has been explored in \cite{Jin2021} and is beyond the scope of this study. By treating $\mathbf{u}$ as known, we decouple the problem and focus exclusively on solving the advection-diffusion equation to model pollutant dispersion.

The forward problem consists of solving the advection-diffusion equation given the pollutant source term $s$, the wind velocity field $\mathbf{u}$, and other parameters such as $k$. The goal is to determine the pollutant concentration $c$ as a function of space and time. Mathematically, the forward problem can be expressed as:
\[
\text{Given } \mathbf{u}, k, \text{ and } s, \text{ solve for } c(x, y, t).
\]

The inverse problem involves determining the unknown source term $s$, which describes the location and intensity of pollutant emissions, using discrete ground-based measurements of the pollutant concentration $c$. The objective is to identify $s$ such that the advection-diffusion equation is satisfied for the observed data. Mathematically, the inverse problem can be formulated as:
\[
\text{Given } c(x, y, t) \text{ (measured data), solve for } s(x, y).
\]

Nondimensionalization of equations is a common procedure to convert them into dimensionless variables. This approach simplifies the analysis of the problem and ensures that all terms in the equation are of the same order of magnitude. As shown in \cite{Chuprov2024,chuprova2024pinn}, nondimensionalization can have a positive impact on the convergence of PINNs. By ensuring that the terms in the equation are balanced, the sensitivity of the loss function becomes equal for all PDE terms, leading to more stable and efficient training.
When scaling the advection-diffusion equation, we use characteristic values for concentration $\mathit{C}$, length $\mathit{L}$, and velocity $\mathit{U}$. The dimensionless variables are defined as follows:

\begin{equation}
\mathit{c}^* = \frac{\mathit{c}}{\mathit{C}}, \quad \mathbf{u}^* = \frac{\mathbf{u}}{\mathit{U}}, \quad \mathit{t}^* = \frac{\mathit{t}\mathit{U}}{\mathit{L}}. \label{normalization}
\end{equation}
Substituting these dimensionless variables into Eq. (\ref{eq1}) yields:
\begin{equation}
   \frac{\partial c^*}{\partial t^*} + \mathbf{u} \cdot \nabla c^* - 
\frac{1}{\text{Pe}}
\nabla^2 c^* = s, 
\end{equation}
where $\text{Pe}$ is the Peclet number, given by
\begin{equation}
\text{Pe} = \frac{\mathit{L}\mathit{U}}{\mathit{k}}.    
\end{equation}

\subsection{PINN methodology}
Here we briefly summarize the principles of PINN. PINN is a machine learning approach that combines data-driven learning with physical laws expressed as PDEs. Unlike traditional neural networks, which rely solely on data for training, PINNs embed governing physical equations directly into their loss function. This ensures that the solutions not only fit the observed data but also satisfy fundamental physical principles.
In this study, we apply the PINN framework to solve 
the advection-diffusion equation (\ref{eq1}). In this case, PINN takes spatial coordinates $(x, y)$ and time $t$ as input and predicts the pollutant concentration $c(x, y, t)$ as output. In this way, PINN approximates the solution to the advection-diffusion equation:
\[
c_\text{PINN}(x, y, t; \theta),
\]
where $\theta$ represents the trainable parameters (weights and biases) of the neural network.

The loss function in a PINN framework is designed to enforce both data consistency and physical law adherence. It consists of three primary components listed below.

1. PDE residual loss ensures that the predicted solution satisfies the advection-diffusion equation:
   \begin{equation}
   \mathcal{L}_\text{PDE} = \frac{1}{N_\text{collocation}} \sum_{i=1}^{N_\text{collocation}} \left| \frac{\partial c_\text{PINN}}{\partial t} + \mathbf{u} \cdot \nabla c_\text{PINN} - \nabla \cdot (k \nabla c_\text{PINN}) - s_\text{PINN} \right|^2.       
   \end{equation}
   Here, $N_\text{collocation}$ represents the number of collocation points in the domain.

2. Data loss minimizes the discrepancy between the predicted and observed pollutant concentrations at ground-based measurement points:
   \begin{equation}
   \mathcal{L}_\text{data} = \frac{1}{N_\text{data}} \sum_{i=1}^{N_\text{data}} \left| c_\text{PINN}(x_i, y_i, t_i; \theta) - c_\text{obs}(x_i, y_i, t_i) \right|^2,
   \end{equation}
   where $c_\text{obs}$ denotes the observed concentration values and $N_\text{data}$ is the number of measurement points.

3. Loss of the boundary and initial condition ensures that the solution satisfies the boundary and initial conditions of the problem:
   \begin{equation}
   \mathcal{L}_\text{BC} = \frac{1}{N_\text{boundary}} \sum_{i=1}^{N_\text{boundary}} \left| c_\text{PINN}(x_i, y_i, t=0) - c_\text{initial}(x_i, y_i) \right|^2.
  \end{equation}

\begin{figure}[t]
    \centering
    \includegraphics[width=0.7\textwidth]{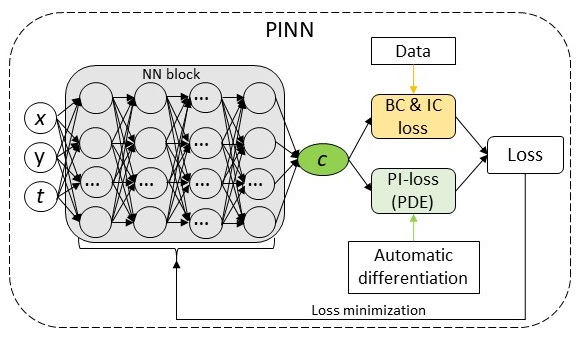}
    \caption{Example of a PINN architecture. The network takes spatial coordinates $(x, y)$ and time $t$ as input and predicts the pollutant concentration $c(x, y, t)$. The loss function combines data consistency, PDE residuals, and boundary conditions.}
    \label{fig:pinn-schematic}
\end{figure}

The total loss function minimized during the training is a weighted combination of these components:
\begin{equation}
\mathcal{L}_\text{total} = \lambda_\text{PDE} \mathcal{L}_\text{PDE} + \lambda_\text{data} \mathcal{L}_\text{data} + \lambda_\text{BC} \mathcal{L}_\text{BC}, \label{loss}
\end{equation}
where $\lambda_\text{PDE}$, $\lambda_\text{data}$, and $\lambda_\text{BC}$ are weights balancing the contributions of each term. Depending on the type and formulation of the problem, certain terms in the equation may not be used. For example, in the forward problem, the term corresponding to data losses $\mathcal{L}_\text{data}$ may be omitted, while in the inverse problem, the term related to boundary conditions $\mathcal{L}_\text{BC}$ may be excluded.
The training process involves minimizing the total loss function $\mathcal{L}_\text{total}$ using a gradient-based optimizer such as Adam or L-BFGS. During training the neural network predicts $c_\text{PINN}$ for collocation points and measurement points. The loss function is evaluated and backpropagated to update the network parameters $\theta$.
The schematic of the PINN architecture is illustrated in Figure \ref{fig:pinn-schematic}.

\section{Implementation of PINN}

We developed a versatile framework for PINN designed to solve a wide range of PDEs, including the advection-diffusion equation. This framework is built on PyTorch. By integrating GPU acceleration, our implementation enables efficient training and large-scale simulations.
The framework is deployed on the high-performance computing (HPC) cluster at HSE University \cite{Kostenetskiy2021}, which provides the necessary computational resources to handle complex simulations and large datasets. The network training was conducted on an NVIDIA V100 32 GB SXM GPU.

It is known that in certain cases, PINNs may fail to converge or produce suboptimal results due to issues such as local minima in the loss function, inefficient parameter optimization, or architectural limitations \cite{Wang2021,Wang2021a}. To address these challenges, various strategies are implemented in the framework to enhance the convergence and accuracy of PINN. Next, we discuss some of the most promising methods.

One way to improve PINN performance is through advanced architectures like First-Order PINN (FO-PINN) \cite{gladstone2023_fopinn}. FO-PINN modifies the standard PINN by using both the function values and their first derivatives to approximate solutions to PDEs. Unlike traditional PINN, which rely on second derivatives calculated using automatic differentiation, FO-PINN evaluates second derivatives indirectly using additional neural network outputs and corresponding loss terms. This approach reduces computational overhead and improves model accuracy.

Another approach to enhancing PINN performance is the Separable PINN (SPINN) \cite{cho2023_spinn}. SPINN represents the target function of multiple variables as a product or sum of functions of single variables. Instead of using collocation points across the entire domain, SPINN employs vectors of points along individual axes. This separation of variables allows the use of simpler neural networks, reduces the number of trainable parameters, and improves computational efficiency.

One common reason for PINN convergence issues is the network getting trapped in local minima of the loss function. A promising strategy to mitigate this issue is to use sinusoidal mapping of input data \cite{Wong2024_sin}. This approach involves applying a sinusoidal activation function in the first layer:
\begin{equation}
\gamma(x) = \sin(2\pi(Wx + b)),
\label{eq:sin_input}
\end{equation}
where \(W\) and \(b\) are trainable parameters. 

The selection of weights in Eq. (\ref{loss}) is another critical factor in improving PINN convergence. An adaptive algorithm for selecting weights, such as the one proposed in \cite{Wang2022}, did not yield significant improvements in our experiments. 
During training, weights $\lambda$ are adjusted to ensure that the terms in Eq. \ref{loss} are of the same order of magnitude. This adjustment not only reduces the residual error in initial conditions but also improves overall accuracy.

\section{Results}
\subsection{Solution of the forward model}

The forward problem involves modeling the distribution of pollutants in a computational domain, given a predefined velocity field. The goal is to use PINN to simulate the advection and diffusion processes and evaluate their accuracy in reproducing pollutant dispersion.
For our applications, the forward problem is considered solved successfully if the mean squared error (MSE) between the PINN-predicted pollutant distribution and the finite element method (FEM) solution remains below \(1 \times 10^{-3}\) for all time steps. FEM is chosen as the reference method due to its well-established accuracy in solving partial differential equations, making it a reliable benchmark for evaluating PINN performance.
Solving the forward problem first provides valuable insights into the model's behavior, simplifying the subsequent solution of the inverse problem.

\begin{figure}[tbh!]
    \centering
    \includegraphics[width=0.8\textwidth]{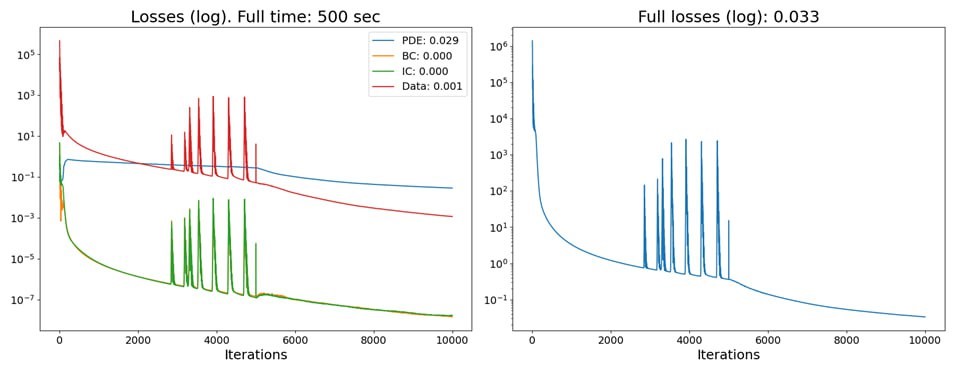}
    \caption{Loss curves of PINN for the advection-diffusion equation}
    \label{fig:ADforward_losses}
\end{figure}

\begin{figure}[tbh!]
    \centering
    \includegraphics[width=0.8\textwidth]{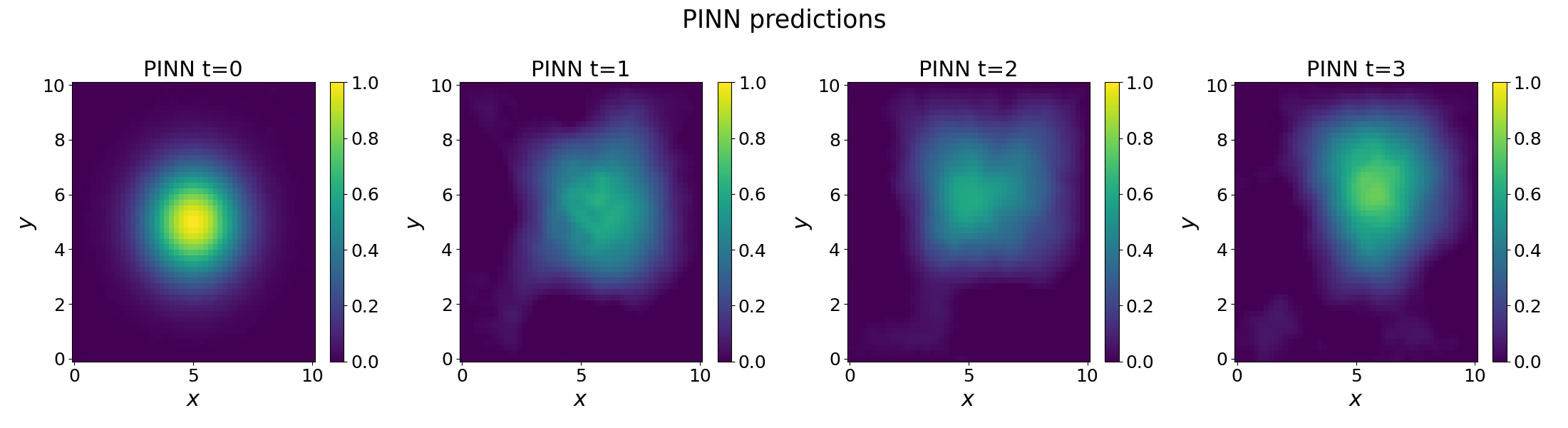}
    \centering
    \includegraphics[width=0.8\textwidth]{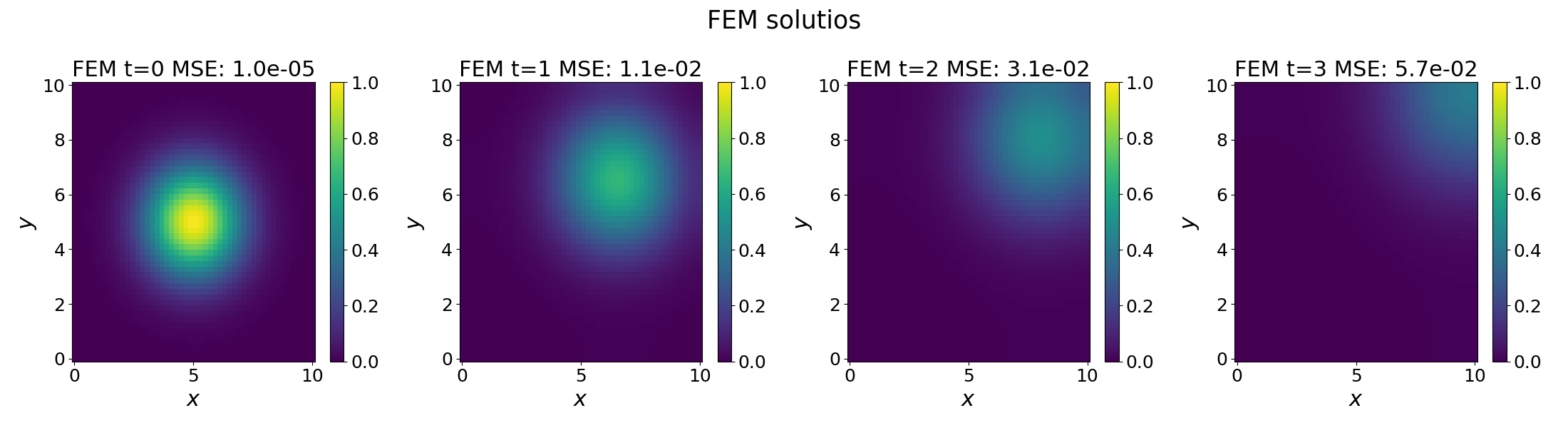}
    \centering
    \includegraphics[width=0.8\textwidth]{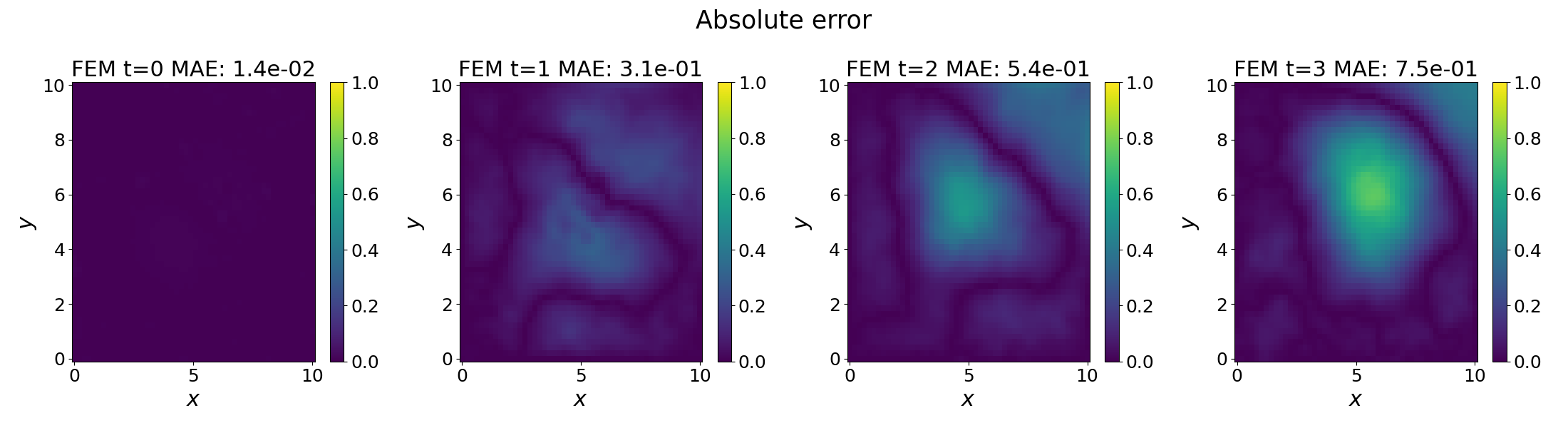}
    \caption{Comparison of classical PINN results with FEM results for the advection-diffusion equation, t = 0-4}
    \label{fig:ADforward_pinn}
\end{figure}

We iteratively optimized the PINN architecture for the advection-diffusion equation. Initially, we tuned the architecture parameters manually; however, for improved results, one could employ stochastic search algorithms as outlined in \cite{Buzaev2023}. The preliminary results for the PINN, which uses fully connected layers with 300 neurons per layer applied to the original equation \ref{eq1}, are presented in Figures \ref{fig:ADforward_losses} and \ref{fig:ADforward_pinn}. Specifically, Figure \ref{fig:ADforward_losses} shows the components of the loss function (left) and the total loss (right). The PDE loss curve levels off, suggesting inadequate training. Meanwhile, Figure \ref{fig:ADforward_pinn} compares the PINN predictions (first row) to the FEM solutions (second row), along with the absolute error (third row) at times \(t = 0, 1, 2,\) and \(3\) seconds.

\begin{figure}[tbh!]
    \centering
    \includegraphics[width=0.8\textwidth]{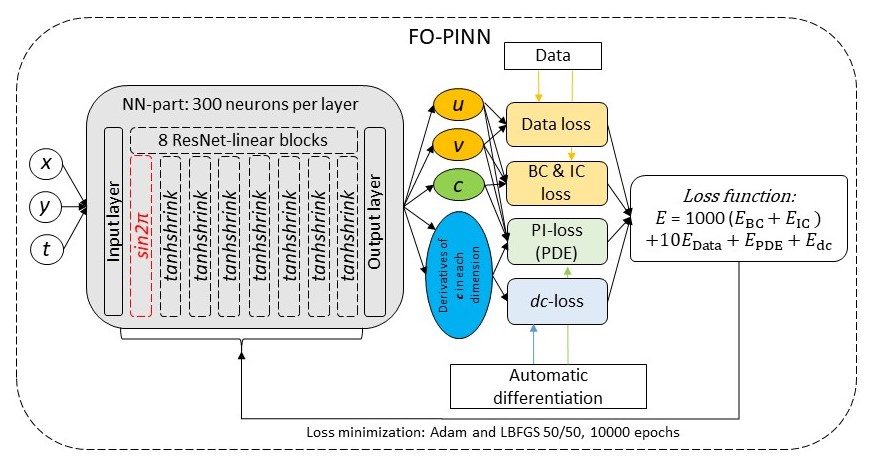}
    \caption{FO-PINN architecture for solving the forward problem of the advection-diffusion equation}
    \label{fig:ADforward_architecture}
\end{figure}

The optimal architecture incorporates First-Order PINN (FO-PINN) with \(\lambda = 1000\) for initial and boundary conditions, together with ResNet blocks. ResNet blocks, which combine linear layers with residual connections, mitigate vanishing gradient issues and improve training stability. Each layer contains 300 neurons. The weighting factor \(\lambda = 1000\) was chosen to prioritize initial and boundary conditions. 
The activation functions used are tanhshrink, with the $\sin{2\pi}$ activation function applied to the first input layer. The optimizers used in this study are Adam and LBFGS, applied in a 50/50 ratio, with the total number of iterations set to 10000. In addition, the nondimensionalization procedure significantly improves the convergence behavior.
The training process takes approximately 8 minutes when run on a GPU.
Figure \ref{fig:ADforward_architecture} illustrates the tuned PINN architecture.

\begin{figure}[tbh!]
    \centering
    \includegraphics[width=0.8\textwidth]{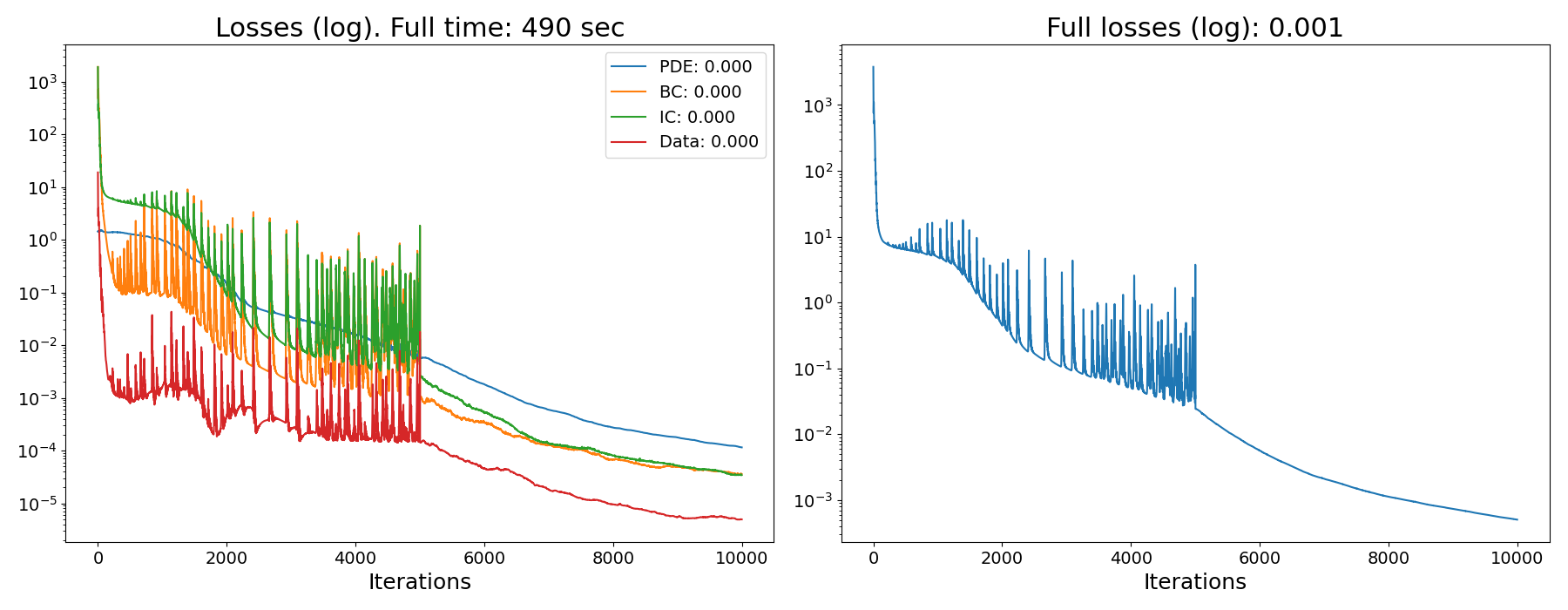}
    \caption{Loss curves of the improved PINN for the normalized advection-diffusion equation.}
    \label{fig:ADforward_losses_final}
\end{figure}

\begin{figure}[tbh!]
    \centering
    \includegraphics[width=0.9\textwidth]{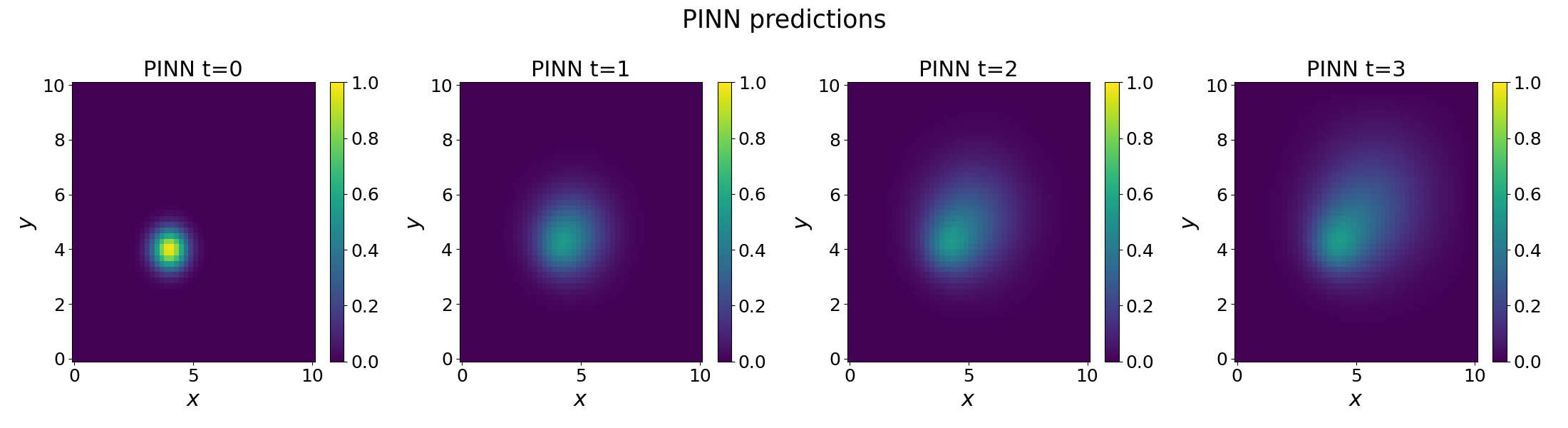}
    \centering
    \includegraphics[width=0.9\textwidth]{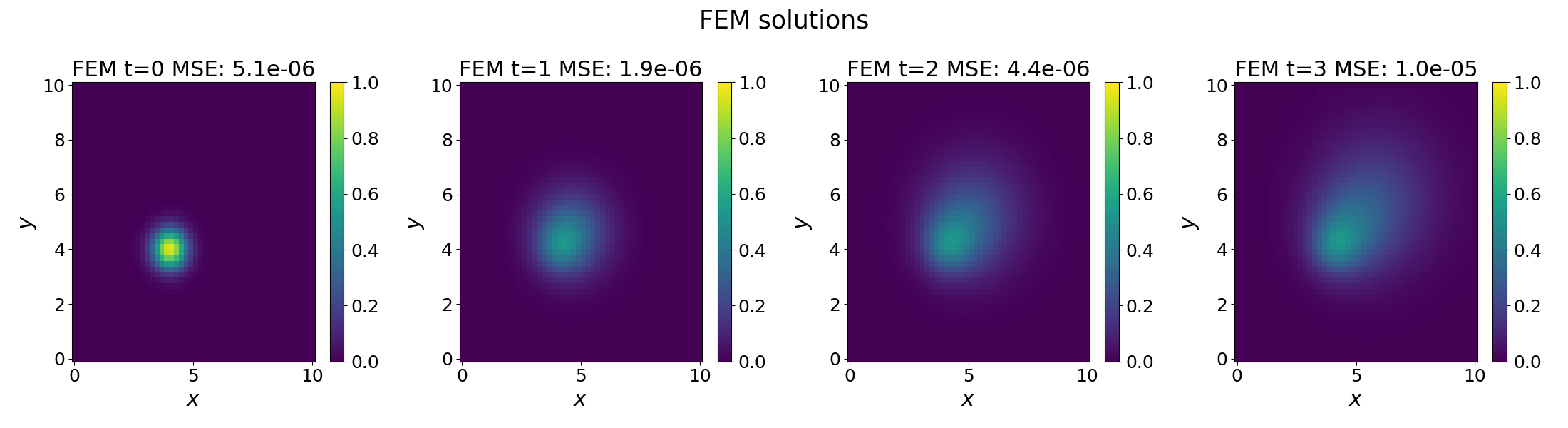}
    \centering
    \includegraphics[width=0.9\textwidth]{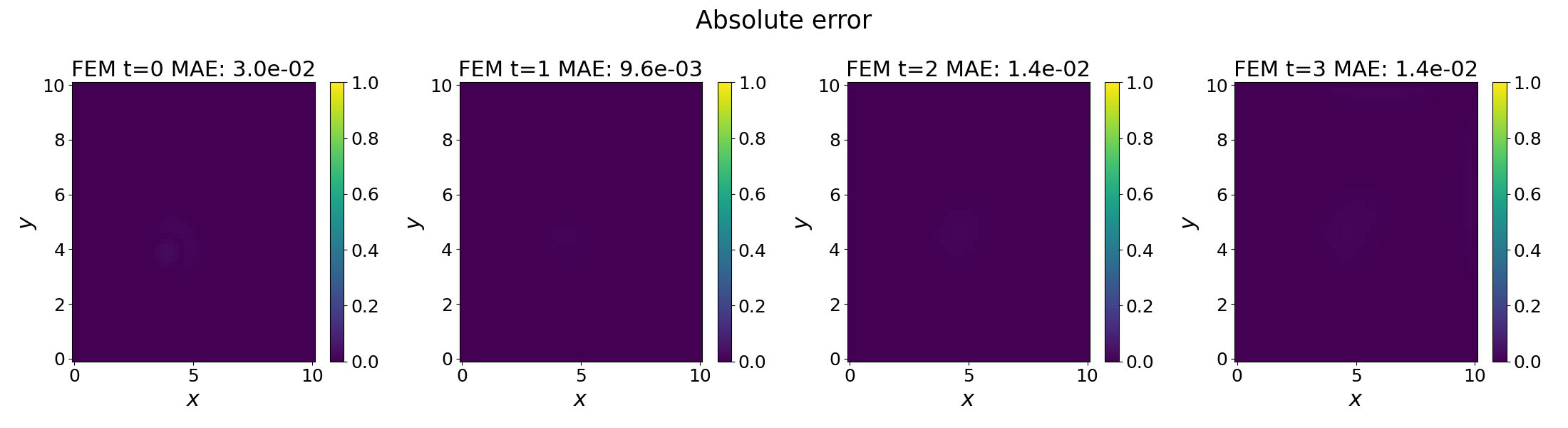}
    \caption{Results for the forward problem of the normalized advection-diffusion equation: improved PINN}
    \label{fig:ADforward_pinn_final}
\end{figure}

Figure \ref{fig:ADforward_losses_final} shows the loss curves, while Figure \ref{fig:ADforward_pinn_final} illustrates the obtained solutions.
The improved PINN achieves good agreement with the FEM results, with MSE values ranging from \(1 \times 10^{-5}\) to \(1 \times 10^{-6}\) and mean absolute error (MAE) values between \(1 \times 10^{-2}\) and \(1 \times 10^{-3}\). These results demonstrate the effectiveness of the optimized PINN configuration in accurately modeling pollutant dispersion. With the optimal PINN setup in place, we now shift our focus to the inverse problem of identifying pollution sources using the same framework.

\subsection{Solution of the inverse problem using synthetic data}

The inverse problem involves determining the location and characteristics of an unknown pollution source based on limited observations of pollutant concentrations at specific points within the computational domain. This includes estimating both the coordinates and possibly the intensity of the source. Solving this problem has significant practical importance, as it enables the identification of pollution sources and the assessment of their environmental impact. In this section, we evaluate the effectiveness of PINN using test data generated by FEM to assess the accuracy and reliability of our model before applying it to real-world data. We define multiple synthetic scenarios, including cases with constant and variable winds, different diffusion coefficients, and varying numbers of sources.

\begin{figure}[tbh!]
    \centering
    \includegraphics[width=0.8\textwidth]{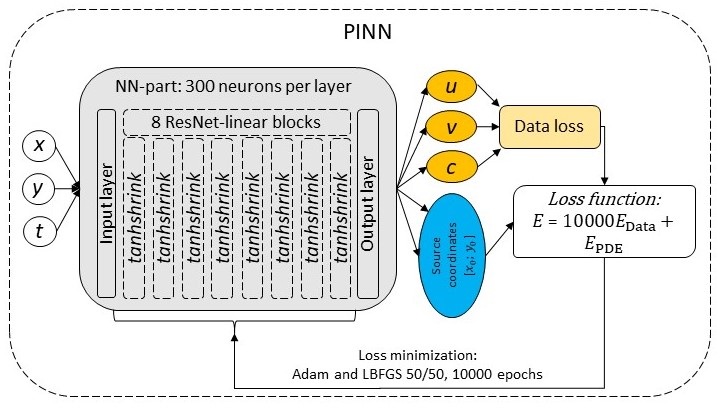}
    \caption{Optimized PINN architecture for solving the inverse problem of the advection-diffusion equation.}
    \label{fig:ADinverse_architecture}
\end{figure}

To effectively address the inverse problem, we introduced several modifications to the PINN architecture. The FO-PINN architecture and the $\sin{2\pi}$ activation function in the first layer were removed as they do not contribute to solving the inverse problem. 
The value of $\lambda_\text{data}$ in Eq. (\ref{loss}) is set to 10000.
The final model configuration optimized for solving the inverse problem of the advection-diffusion equation is illustrated in Figure \ref{fig:ADinverse_architecture}.

To identify the unknown source, we treat its coordinates as trainable parameters within the neural network. Similar to weights and biases, these parameters are iteratively updated using gradient-based optimization during the training process. An alternative approach would be to include additional outputs in the neural network and introduce corresponding loss function terms for these outputs. However, for the problem considered, this approach yielded unsatisfactory results and therefore not considered in this paper.
In our simulations, the coordinates of the pollution source are initialized either as a randomly selected point within the computational domain or as its midpoint. During training, PINN substitutes these parameters into the governing equation and progressively refines the estimated source location.

\begin{figure}[tbh!]
    \centering
    \includegraphics[width=0.8\textwidth]{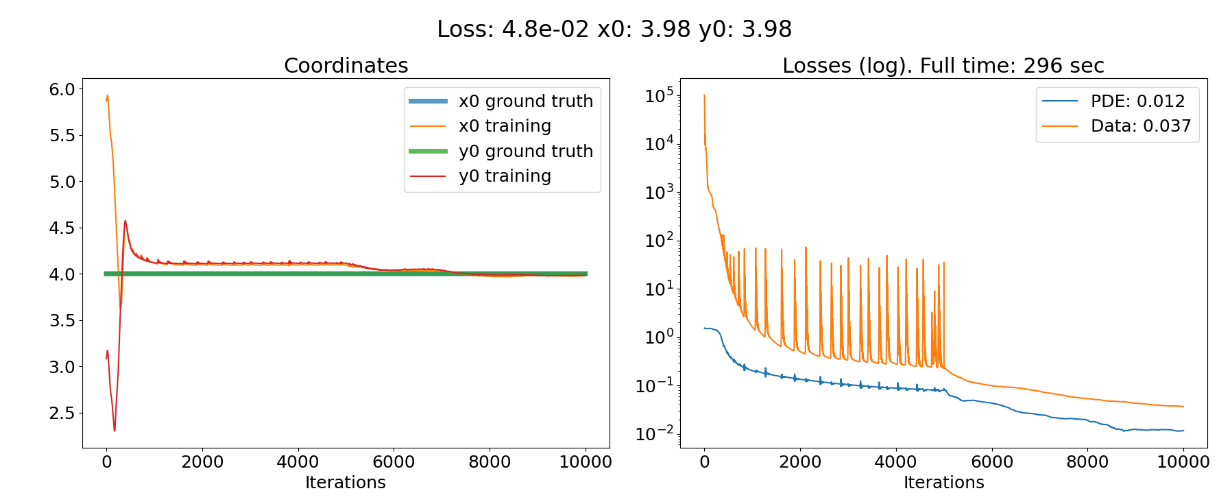}
    \caption{The loss function plot and the process of locating the coordinates for t = 3 and constant wind speed}
    \label{fig:ADinverse_losses}
\end{figure}

\begin{figure}[tbh!]
    \centering
    \includegraphics[width=0.8\textwidth]{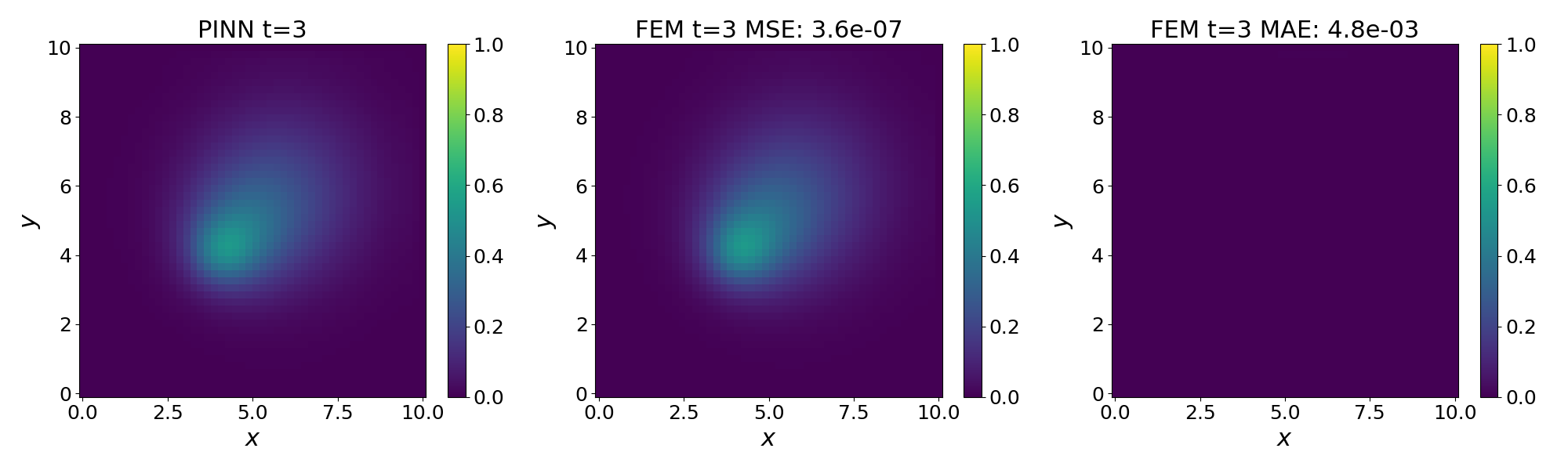}
    \caption{Comparison of PINN results with FEM data for t = 3 and constant wind speed}
    \label{fig:ADinverse_results}
\end{figure}

In the first scenario, we place a point source at coordinates \( (4,4) \) in the domain \( (10,10) \) with a diffusion coefficient of \( k = 0.5 \), constant wind speeds \( u = 0.7 \) and \( v = 0.7 \), and consider a single time instance at \( t = 3 \). The Peclet number for this case is 11.8.
Figure \ref{fig:ADinverse_losses} presents the loss function plots and how the estimated coordinates converge towards the target source coordinates during training. In Figure \ref{fig:ADinverse_results}, we present the reference solution obtained using FEM alongside the PINN solution with the identified pollution source coordinates, demonstrating a good agreement between the two.

Table \ref{tab:ADinverse_results} provides a comparison between the predicted source coordinates and the target values, along with the corresponding MSE.
During the study, we observed that the number of training epochs had a more significant impact on the results than the number of collocation points. Our experiments were conducted for 10000–15000 epochs and 200–5000 collocation points. The results obtained for 15000 epochs with only 200 collocation points were comparable to those obtained with 10000 epochs and 2500 collocation points. The detailed comparison is presented in Table \ref{tab:ADinverse_3cases}. This finding is particularly relevant for future studies, where real-world data availability may be limited to a small number of sensors.

\begin{table}[tbh!]
	\centering
	\caption{Comparison of true source coordinates with predicted coordinates using PINN for t = 3 and constant wind speed}
	\begin{tabular}{|c|c|c|c|}
 		\hline
 		& \textbf{True coordinates} & \textbf{Predicted coordinates} & \textbf{MSE} \\ \hline
 		\textbf{x} & 4.0 & 3.979 & 4.4e-4 \\ \hline
 		\textbf{y} & 4.0 & 3.982 & 3.2e-4 \\ \hline
 	\end{tabular}
 	\label{tab:ADinverse_results}
\end{table}

\begin{table}[tbh!]
	\centering
	\caption{Predicted coordinates for three cases: 1) 10000 epochs, 2500 collocation points; 2) 15000 epochs, 2500 collocation points; 3) 15000 epochs, 200 collocation points}
	\resizebox{\textwidth}{!}{
	\begin{tabular}{|c|c|c|c|c|c|c|c|c|c|}
		\hline
		& \multicolumn{3}{|c|}{\textbf{1 case}} & \multicolumn{3}{|c|}{\textbf{2 case}} & \multicolumn{3}{|c|}{\textbf{3 case}} \\ \hline
 		\hline
 		& \parbox{2cm}{\textbf{\centering True \\ coordinates}} & \parbox{2cm}{\textbf{\centering Predicted \\ coordinates}} & \textbf{MSE} & \parbox{2cm}{\textbf{\centering True \\ coordinates}} & \parbox{2cm}{\textbf{\centering Predicted \\ coordinates}} & \textbf{MSE} & \parbox{2cm}{\textbf{\centering True \\ coordinates}} & \parbox{2cm}{\textbf{\centering Predicted \\ coordinates}} & \textbf{MSE} \\ \hline 
 		\textbf{x} & 4.0 & 4.119 & 1.4e-2 & 4.0 & 4.060 & 3.5e-3 & 4.0 & 4.130 & 1.7e-2\\ \hline
 		\textbf{y} & 4.0 & 4.091 & 8.2e-3 & 4.0 & 4.085 & 7.2e-3 & 4.0 & 4.099 & 9.7e-3 \\ \hline
 	\end{tabular}}
 	\label{tab:ADinverse_3cases}
\end{table}

\begin{figure}[tbh!]
    \centering
    \includegraphics[width=0.8\textwidth]{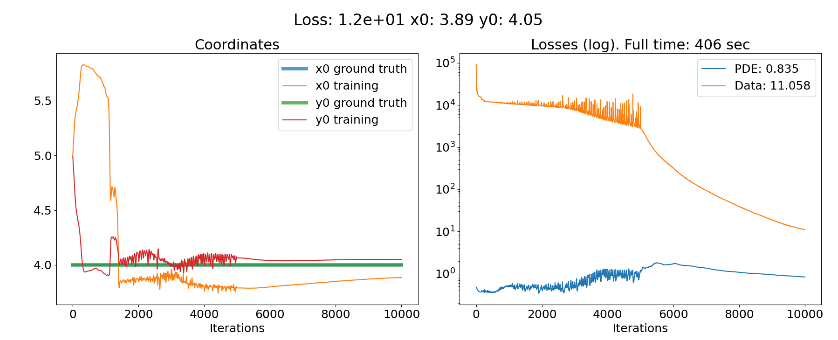}
    \caption{The loss function plot and the process of locating the coordinates for variable wind speed, t = 0-4, and diffusion coefficient k = 0.5}
    \label{fig:ADinverse_losses_t_0_4}
\end{figure}

\begin{figure}[tbh!]
    \centering
    \includegraphics[width=0.9\textwidth]{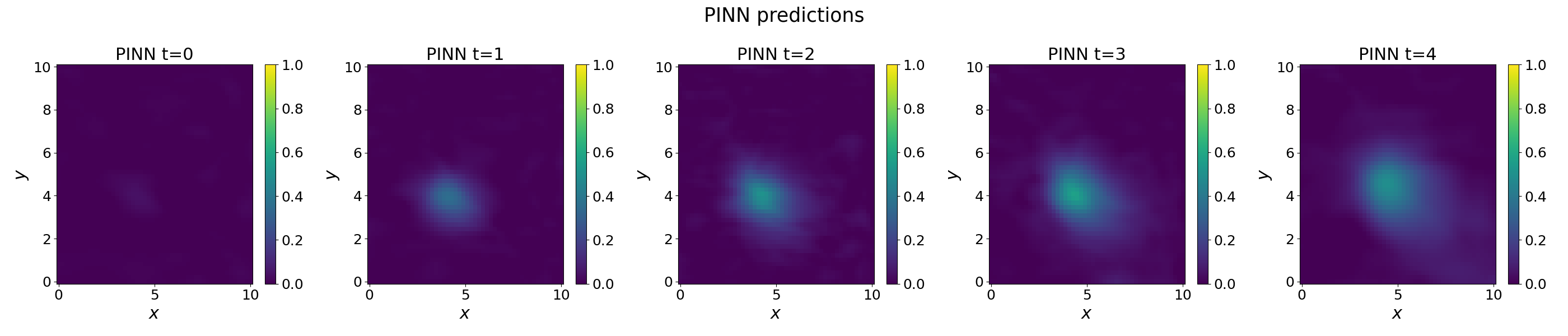}
    \centering
    \includegraphics[width=0.9\textwidth]{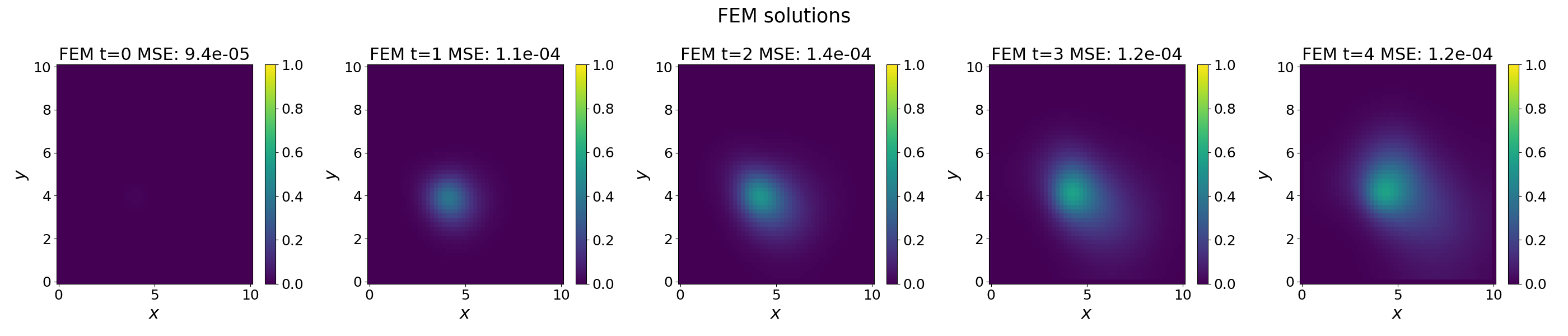}
    \centering
    \includegraphics[width=0.9\textwidth]{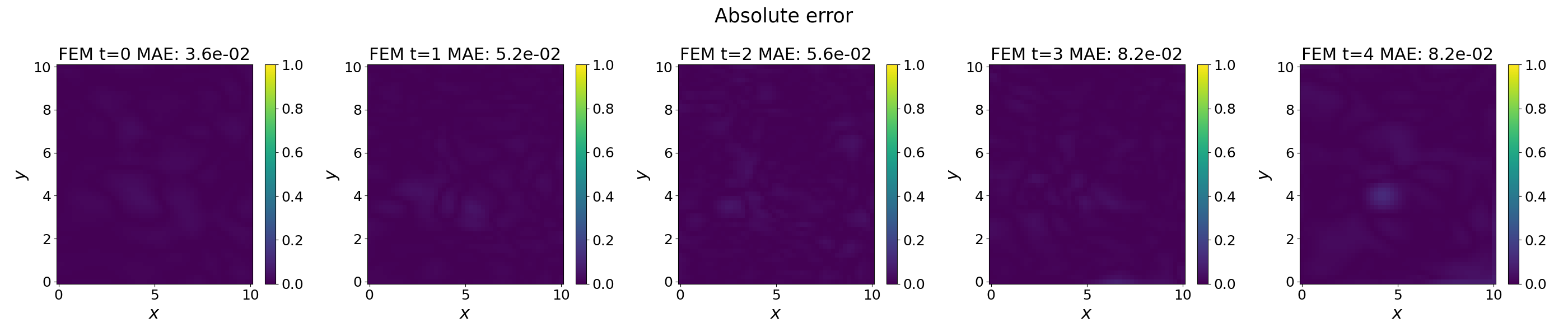}
    \caption{Comparison of PINN results with FEM data for variable wind speed, t = 0-4 seconds, and diffusion coefficient k = 0.5}
    \label{fig:ADinverse_results_t_0_4}
\end{figure}

Next, we consider a more complex case involving variable wind speed field within the time range from 0 and 4. Figure \ref{fig:ADinverse_losses_t_0_4} shows the loss function and evolution of source coordinates. This wind speed field is used throughout the paper unless stated otherwise.
Figure \ref{fig:ADinverse_results_t_0_4} shows a comparison between the results of PINN and the FEM data used for training the network. 
As we can observe, compared to the previous case involving a single time instance, the current scenario presents some challenges for PINN. However PINN still accurately predicts the source coordinates, as shown in Table \ref{tab:ADinverse_results_t_0_4}.

\begin{table}[tbh!] 
	\centering 
	\caption {Comparison of true source coordinates with predicted by PINN for variable wind speed, t = 0-4 seconds, and diffusion coefficient k = 0.5}
	\begin{tabular}{|c|c|c|c|}
 		\hline
 		& \textbf{True coordinates} & \textbf{Predicted coordinates} & \textbf{MSE} \\ \hline
 		\textbf{x} & 4.0 & 3.885 & 1.3e-2 \\ \hline
 		\textbf{y} & 4.0 & 4.049 & 2.4e-3 \\ \hline
 	\end{tabular}
 	\label{tab:ADinverse_results_t_0_4}
\end{table}

\begin{figure}[tbh!]
    \centering
    \includegraphics[width=0.8\textwidth]{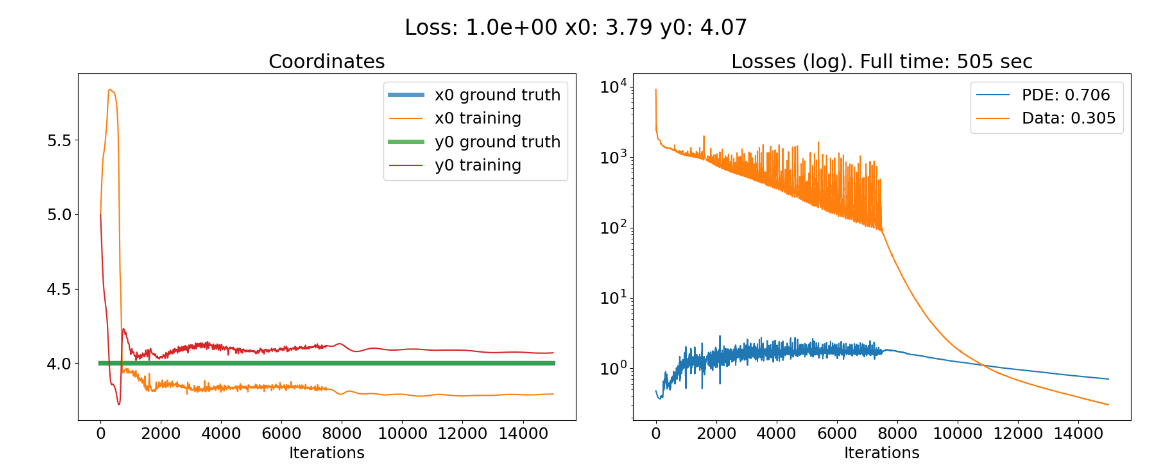}
    \caption{The loss function plot and the process of locating the coordinates for variable wind speed, t = 0-4 seconds, and diffusion coefficient k = 1e-5}
    \label{fig:ADinverse_losses_k1e-5}
\end{figure}

\begin{figure}[tbh!]
    \centering
    \includegraphics[width=0.9\textwidth]{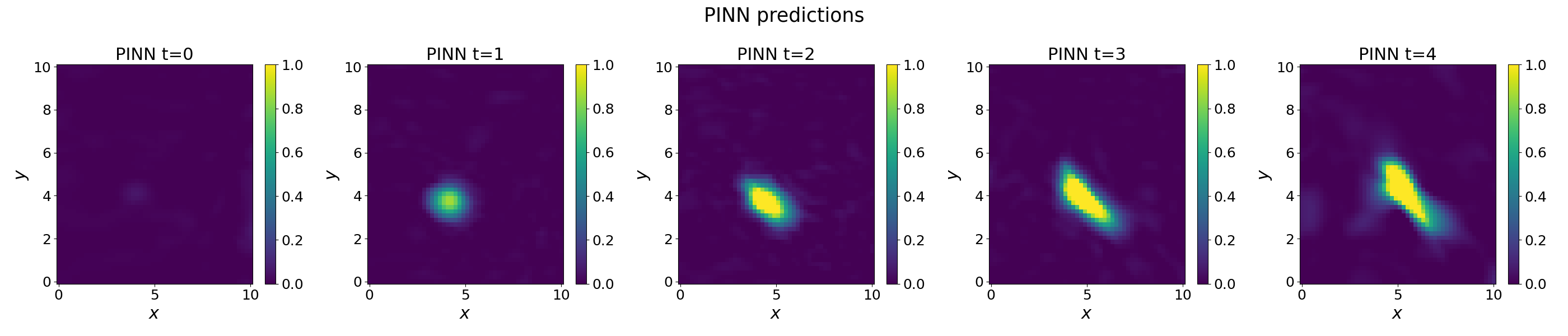}
    \centering
    \includegraphics[width=0.9\textwidth]{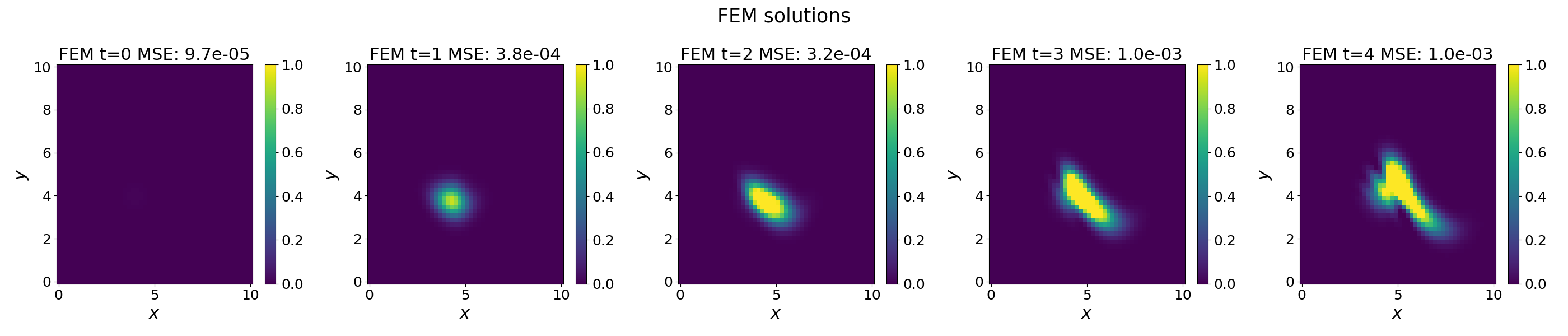}
    \centering
    \includegraphics[width=0.9\textwidth]{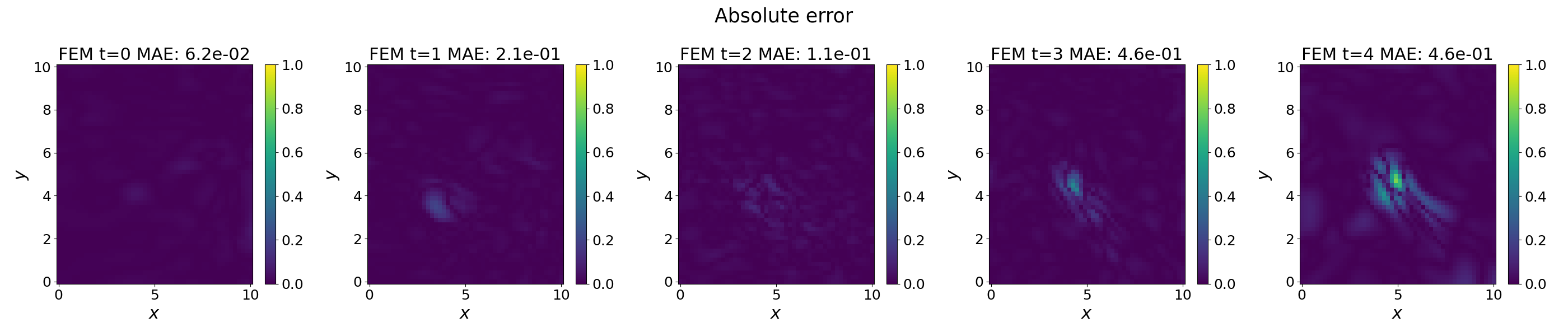}
    \caption{Comparison of PINN results with FEM data for variable wind speed, t = 0-4 seconds, and diffusion coefficient k = 1e-5}
    \label{fig:ADinverse_results_k1e-5}
\end{figure}

Next, we consider a similar scenario but with a diffusion coefficient of \( 10^{-5} \), with the corresponding Peclet number value of 590000. The process of locating the source coordinates and the training plot are presented in Figure \ref{fig:ADinverse_losses_k1e-5}. As the diffusion coefficient decreases, the transport of the substance by the wind becomes more pronounced, slightly impacting the performance of PINN in terms of data learning (Figure \ref{fig:ADinverse_results_k1e-5}) and coordinate prediction (Table \ref{tab:ADinverse_results_k1e-5}). Despite these challenges, PINN continues to exhibit high efficiency in accurately determining the source coordinates.

\begin{table}[tbh!] 
	\centering
	\caption {Comparison of the true source coordinates with the ones predicted by PINN for variable wind speed, t = 0-4 sec, and diffusion coefficient k = 1e-5}
	\begin{tabular}{|c|c|c|c|}
 		\hline
 		& \textbf{True coordinates} & \textbf{Predicted coordinates} & \textbf{MSE} \\ \hline
 		\textbf{x} & 4.0 & 3.793 & 4.3e-2 \\ \hline
 		\textbf{y} & 4.0 & 4.070 & 4.8e-3 \\ \hline
 	\end{tabular}
 	\label{tab:ADinverse_results_k1e-5}
\end{table}

\begin{figure}[tbh!]
    \centering
    \includegraphics[width=0.8\textwidth]{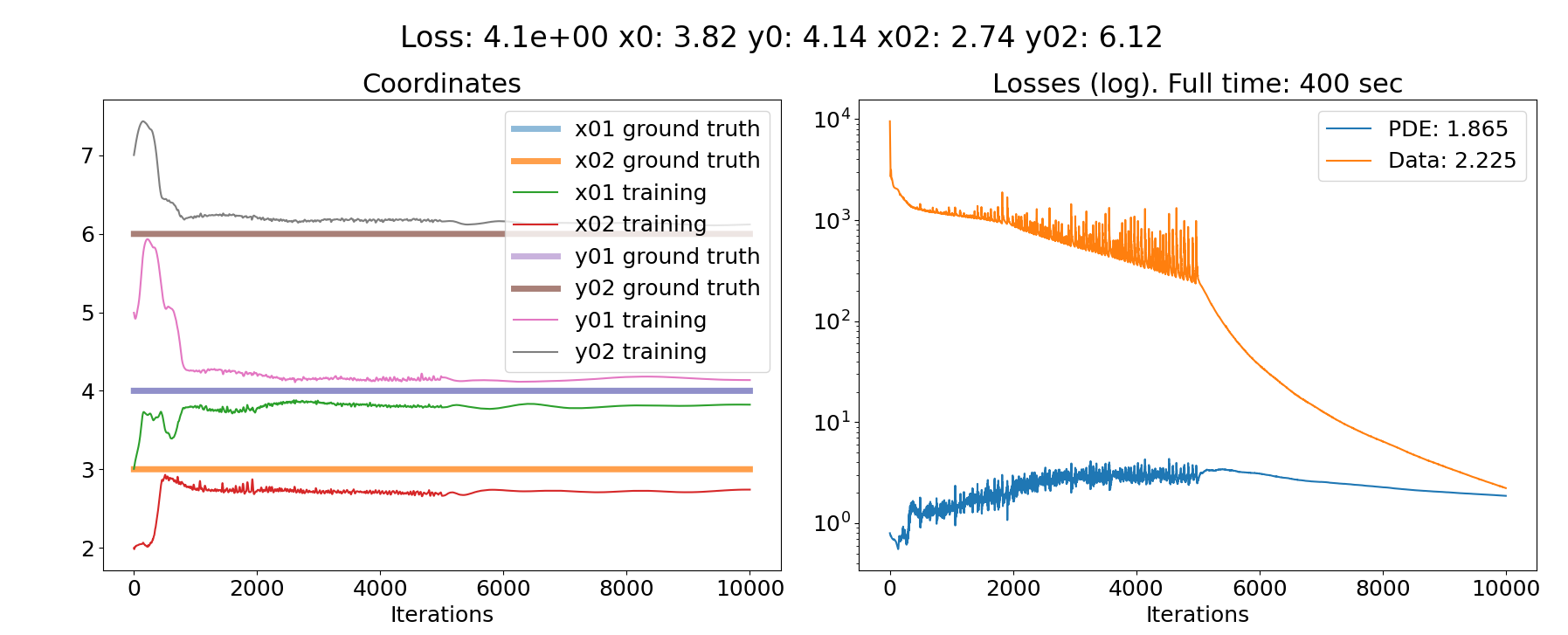}
    \caption{The loss function plot and the process of locating the coordinates for two sources, variable wind speed, t = 0-4, and diffusion coefficient k = 1e-5}
    \label{fig:ADinverse_losses_2_sources}
\end{figure}

\begin{figure}[tbh!]
    \centering
    \includegraphics[width=0.9\textwidth]{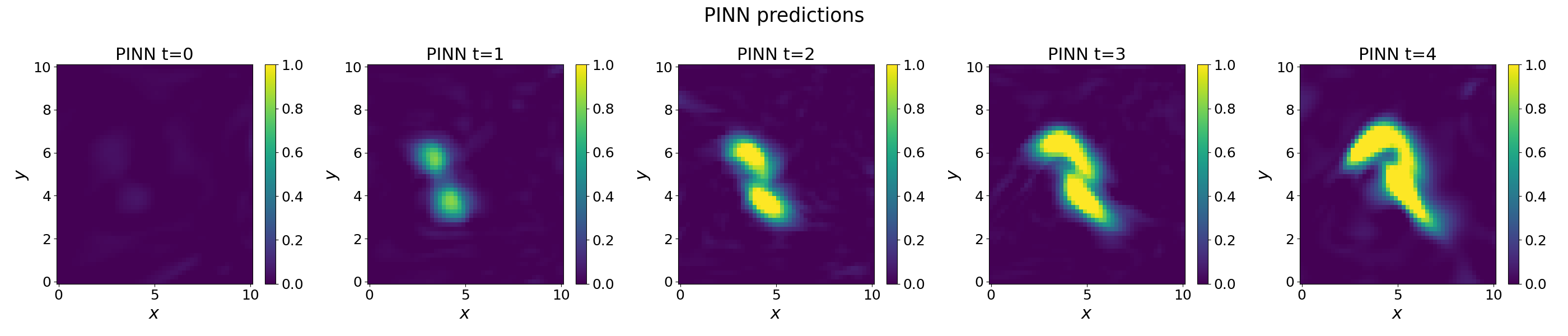}
    \centering
    \includegraphics[width=0.9\textwidth]{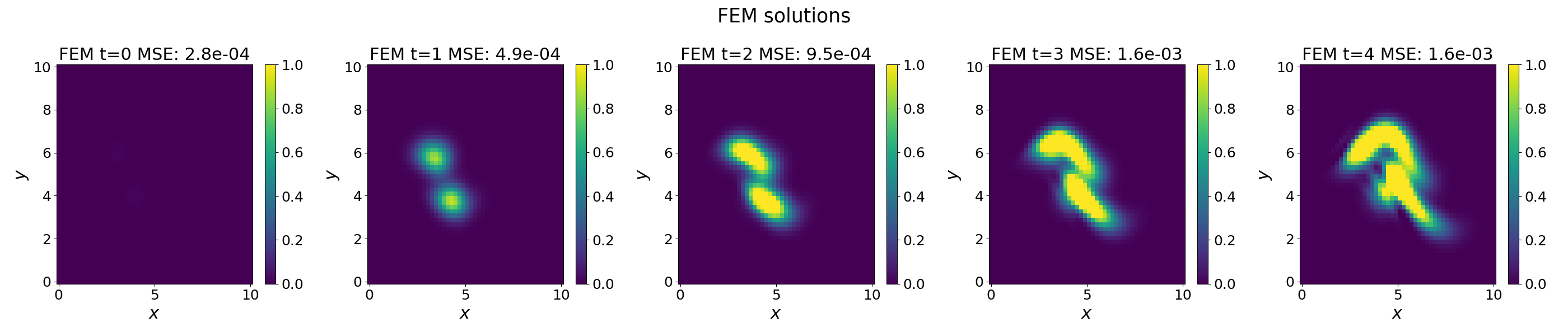}
    \centering
    \includegraphics[width=0.9\textwidth]{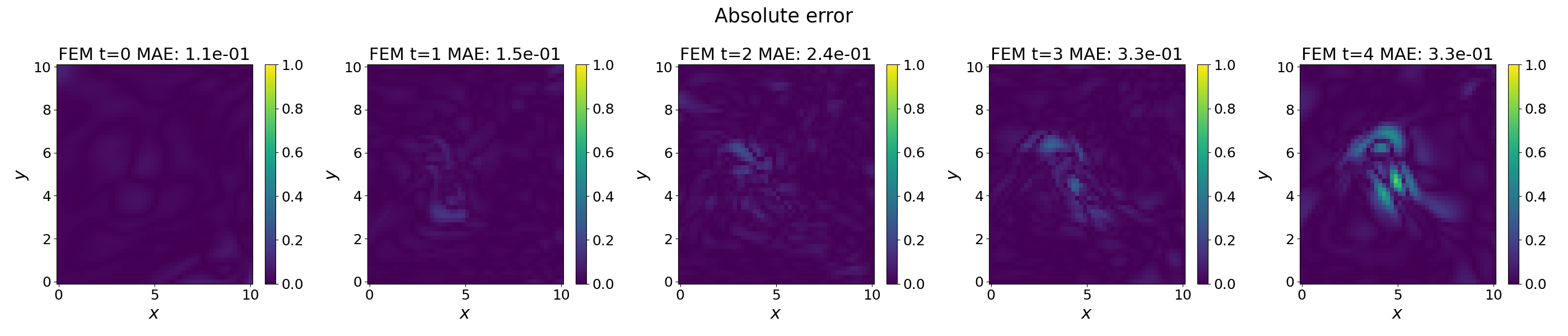}
    \caption{Comparison of PINN results with FEM data for two sources, variable wind speed, t = 0-4, and diffusion coefficient k = 1e-5}
    \label{fig:ADinverse_results_2_sources}
\end{figure}

Next, we consider a scenario similar to the previous one, but with two sources located at coordinates \((4,4)\) and \((3,6)\). Figure \ref{fig:ADinverse_losses_2_sources} illustrates the process of identifying the source coordinates along with the corresponding loss function. Figure \ref{fig:ADinverse_results_2_sources} presents the PINN predictions and their comparison with the reference data. Table \ref{tab:ADinverse_results_2_sources} provides a quantitative comparison between the predicted and true source coordinates.

\begin{table}[tbh!] 
	\centering
	\caption{Comparison of the target coordinates of two sources with the predicted coordinates by PINN for variable wind speed, t = 0-4 sec, and diffusion coefficient k = 1e-5}
	\begin{tabular}{|c|c|c|c|}
 		\hline
 		& \textbf{True coordinates} & \textbf{Predicted coordinates} & \textbf{MSE} \\ \hline
 		\textbf{x1} & 4.0 & 3.823 & 3.1e-2 \\ \hline
 		\textbf{y1} & 4.0 & 4.137 & 1.9e-2 \\ \hline
 		\textbf{x2} & 3.0 & 2.740 & 6.7e-2 \\ \hline
 		\textbf{y2} & 6.0 & 6.121 & 1.5e-2 \\ \hline
 	\end{tabular}
 	\label{tab:ADinverse_results_2_sources}
\end{table}

To validate the proposed approach, we apply PINN to a set of test pollution source coordinates within the domain \((10,10)\): \((3,3)\), \((4,4)\), \((5,5)\), \((6,6)\), \((7,7)\), \((4,5)\), \((5,2)\), \((6,3)\), \((4,6)\), and \((2,4)\). 
To assess the accuracy of the predicted source coordinates, we compare the coordinates obtained using the PINN-based algorithm with the test coordinates using the MSE metric. 
Table \ref{tab:verification} presents a comparison between the pollution source coordinates predicted by the PINN algorithm and the reference test coordinates. The results indicate that PINN successfully identifies the source locations in over 80\% of the cases, with the error for any correctly predicted coordinate component not exceeding \( 10^{-1} \). 
We obtained robust results across various scenarios, including simple and complex wind fields, different diffusion coefficient values, and cases with both single and multiple sources. Additionally, we tested the PINN method on multiple pollution source coordinates, further confirming its reliability and adaptability. These findings demonstrate the flexibility and accuracy of the proposed approach. 

\begin{table}[tbh!]
    \centering
    \caption{Verification results for different test coordinates}
    \renewcommand{\arraystretch}{1.2}
    \begin{tabular}{|c|c|c|c|c|}
        \hline
        \textbf{№} & \textbf{Test Coordinates} & \textbf{Predicted Coordinates} & \textbf{MSE} & \textbf{Result} \\ \hline
        1  & (3,3)  & (2.938, 3.193) & (3.9e-3, 3.7e-2) & \ding{51} \\ \hline
        2  & (4,4)  & (3.793, 4.072) & (4.3e-2, 5.2e-3) & \ding{51} \\ \hline
        3  & (5,5)  & (4.881, 4.845) & (1.4e-2, 2.4e-2) & \ding{51} \\ \hline
        4  & (6,6)  & (5.892, 5.639) & (1.2e-2, 1.3e-1) & \ding{51} \\ \hline
        5  & (7,7)  & (5.855, 3.929) & (1.3, 9.4)       & \ding{55} \\ \hline
        6  & (4,5)  & (3.813, 5.167) & (3.5e-2, 2.8e-2) & \ding{51} \\ \hline
        7  & (5,2)  & (5.072, 1.865) & (5.2e-3, 2.1e-2) & \ding{51} \\ \hline
        8  & (6,3)  & (5.909, 2.801) & (8.3e-3, 4.0e-2) & \ding{51} \\ \hline
        9  & (4,6)  & (3.813, 6.103) & (3.5e-2, 1.1e-2) & \ding{51} \\ \hline
        10 & (2,4)  & (5.849, 4.010) & (15, 9.3e-5)     & \ding{51} / \ding{55} \\ \hline
    \end{tabular}
    \label{tab:verification}
\end{table}

\subsection{Solution of the inverse problem using real data}

In this section, we address the solution of the inverse advection-diffusion problem to determine the coordinates of the substance source, relying on real-world observation data provided by our industrial partner, which includes data on hydrogen sulfide concentration as well as wind speed and direction \cite{cityair2023}. 
The real-world data represents a domain of 20 by 20 km, divided into a 100 by 100 grid with a 200-meter step. In this grid, the target source is located at the coordinate (37,44).
To achieve satisfactory results, we increase the number of epochs and slightly adjust the ratio of the Adam and LBFGS optimizers: 10000 epochs for Adam and 3000 epochs for LBFGS.
To ensure the comparability of MSE values calculated from different coordinate types (test and real), the nondimensionalization procedure is applied. This approach helps to maintain the accuracy criterion of 1e-1, which is important when comparing model performance across different scales. 

\begin{figure}[tbh!]
    \centering
    \includegraphics[width=0.8\textwidth]{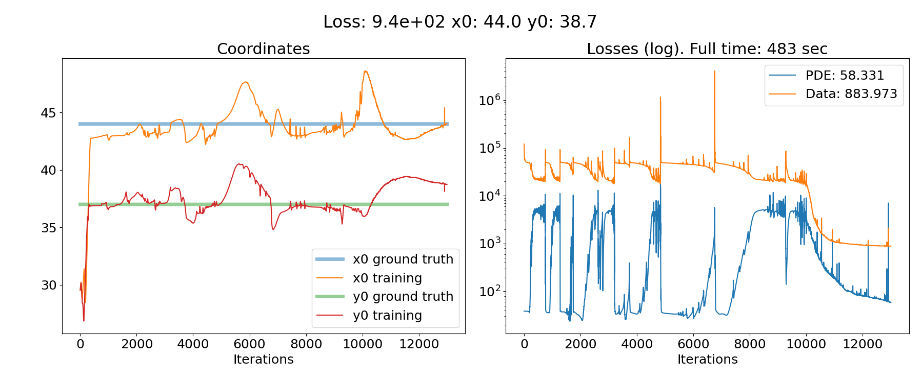}
    \caption{Loss curves and coordinate search process for solving the inverse advection-diffusion problem with real-world data}
    \label{fig:ADinverse_losses_realdata}
\end{figure}

\begin{figure}[tbh!]
    \centering
    \includegraphics[width=0.9\textwidth]{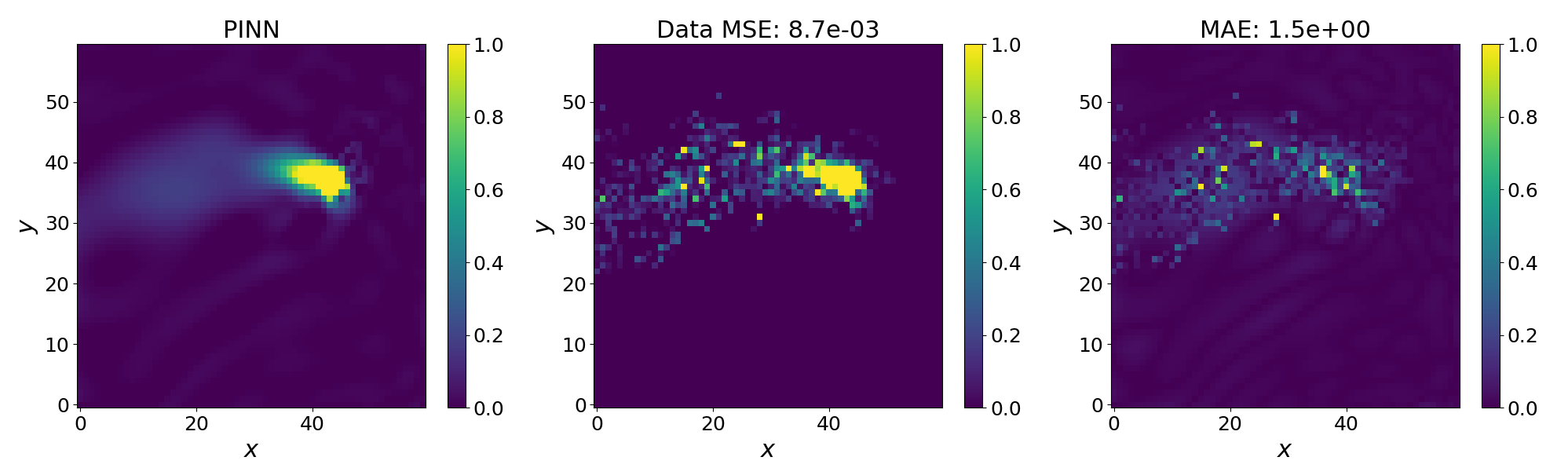}
    \caption{Comparison of PINN results with FEM data for real-world data. The maximum substance concentration value was limited to 1, ensuring better clarity for the analysis}
    \label{fig:ADinverse_results_realdata}
\end{figure}

Results of solving the inverse advection-diffusion problem using real data are presented. Figure \ref{fig:ADinverse_losses_realdata} shows that the approximate coordinates are determined in 500 epochs, which takes less than 20 seconds. However, to achieve optimal accuracy, the increased training time may not have been sufficient. Increasing the number of LBFGS epochs to 5000 resulted in NaN values in the loss function, underscoring the importance of balancing training time and network stability to avoid overfitting or divergence during the optimization process.
Figure \ref{fig:ADinverse_results_realdata} demonstrates how PINN adapts to the real data. While some uncertainty is present in the results obtained by PINN, it does not hinder its ability to locate the source coordinates, which are sufficiently close to the target values, as shown in Table \ref{tab:ADinverse_results_realdata}.
As a result of the conducted research, we observe that PINN method successfully solves inverse advection-diffusion problems, demonstrating high accuracy in determining the source coordinates based on real data. This confirms its effectiveness and potential for further application in this field.

\begin{table}[tbh!] 
	\centering
	\caption{Comparison of true source coordinates with predicted coordinates using PINN}
	\begin{tabular}{|c|c|c|c|}
 		\hline
 		& \textbf{True coordinates} & \textbf{Predicted coordinates} & \textbf{MSE} \\ \hline
 		\textbf{x} & 44.0 & 44.018 & 3.2e-6 \\ \hline
 		\textbf{y} & 37.0 & 38.379 & 1.9e-2 \\ \hline
 	\end{tabular}
 	\label{tab:ADinverse_results_realdata}
\end{table}

\section{Conclusions}
In this study, we have explored the application of PINNs for solving the inverse advection-diffusion problem to localize pollution sources. The PINN solutions have been validated against reference solutions obtained using FEM. To extend the PINN framework to inverse problems, source coordinates were incorporated as trainable parameters within the network.

The proposed approach successfully identifies pollution source coordinates using real-world concentration and wind data, demonstrating its applicability for environmental monitoring. The accuracy of the method depends on the availability of pollutant concentration measurements and wind field data, which can be obtained from meteorological observations or by solving the Navier-Stokes equations. Although this study does not address solving the forward Navier-Stokes problem, future research may focus on integrating PINNs for simultaneous flow and pollutant dispersion modeling.

Despite their potential, PINNs require careful tuning of architecture and hyperparameters, making their application problem-specific and computationally intensive. While in this work, the optimization of the PINN architecture has been performed manually, automating this process through systematic guidelines and global optimization techniques could enhance efficiency and scalability.
Future work will focus on expanding the validation of the approach across diverse atmospheric conditions and pollutant scenarios. Enhancing model accuracy while reducing computational costs remains a key objective. Additionally, transitioning from two-dimensional to three-dimensional modeling will enable more realistic simulations, incorporating factors such as altitude and terrain effects, which are critical for accurate pollutant dispersion modeling.

\section*{Acknowledgments}This research was supported in part through computational resources of HPC facilities at HSE University.

%
%
\bibliographystyle{unsrt}
\bibliography{example-bibtex}{}

\end{document}